\documentclass[letterpaper]{article} 
\usepackage{aaai23}
\usepackage{times}  
\usepackage{helvet}  
\usepackage{courier}  
\usepackage[hyphens]{url}  
\usepackage{graphicx} 
\usepackage{natbib}  
\usepackage{caption} 
\usepackage{algorithm}
\usepackage{algorithmic}

\usepackage{newfloat}
\usepackage{listings}
\DeclareCaptionStyle{ruled}{labelfont=normalfont,labelsep=colon,strut=off} 
\floatstyle{ruled}
\newfloat{listing}{tb}{lst}{}
\floatname{listing}{Listing}
\graphicspath{{./images/}}
\usepackage{xcolor}
\usepackage{subcaption}

\title{Design of Unmanned Air Vehicles Using Transformer Surrogate Models}
\author{
    Adam D. Cobb, Anirban Roy, Daniel Elenius, Susmit Jha
}
\affiliations{

Computer Science Laboratory, SRI International\\
\{adam.cobb, anirban.roy, daniel.elenius, susmit.jha\}@sri.com
}

\usepackage{bibentry}

\begin{document}

\maketitle






\begin{abstract}
Computer-aided design (CAD) is a promising new area for the application of artificial intelligence (AI) and machine learning (ML). The current practice of design of cyber-physical systems uses the {\it digital twin} methodology, wherein the actual physical design is preceded by building detailed models that can be evaluated by physics simulation models. These physics models are often slow and the manual design process often relies on exploring near-by variations of existing designs. AI holds the promise of breaking these design silos and increasing the diversity and performance of designs by accelerating the exploration of the design space. In this paper, we focus on the design of electrical unmanned aerial vehicles (UAVs). The high-density batteries and purely electrical propulsion systems have disrupted the space of UAV design, making this domain an ideal target for AI-based design. In this paper, we develop an AI Designer that 
synthesizes novel UAV designs. Our approach uses a deep transformer model with a novel domain-specific encoding such that we can evaluate the performance of new proposed designs without running expensive flight dynamics models and CAD tools. We demonstrate that our approach significantly reduces the overall compute requirements for the design process and accelerates the design space exploration. Finally, we identify future research directions to achieve full-scale deployment of AI-assisted CAD for UAVs.
\end{abstract}






\section{Introduction}
Artificial intelligence (AI) is transforming multiple application domains, especially those pertaining to areas of creativity and design. Examples of this success can be seen in both art and music ~\cite{briot2020deep,cetinic2022understanding,ramesh2021zero,razavi2019generating}, with particularly impressive results in text-to-image generation \cite{ramesh2022hierarchical,saharia2022photorealistic}, showing the potential of human-AI interaction for creative design. However, in comparison to domains such as text-to-image generation, in scientific design contexts (such as unmanned air vehicles and cyber-physical systems, in general) the requirements of the final design are much more structured and subject to the satisfaction of correctness requirements and optimization of performance objectives. These domains include program synthesis, drug discovery, architecture, and the design of mechanical components~\cite{parisotto2016neuro,korshunova2021openchem,buonamici2020generative,kolata2021decline,granados2021machine}. Unlike the generation of images and text, in these structured-sensitive domains, even generating valid artifacts is nontrivial.

Our paper focuses on the application of machine learning for the design of UAVs. We propose a tool that both speeds up the design process and enables the discovery of new design regimes. One key challenge is in the representation of designs. How can we ensure that the entire UAV design can be captured in a representation that is easily read by any potential machine learning algorithm? A further key challenge is then selecting an appropriate machine learning model. In our work, we show that representing a UAV design as a sequence of tokens built from a preorder traversal of a design tree is sufficient for task-specific design objectives, such as estimating the potential flight distance and hover time of each design. In more detail, we show that representing topological design information as a sequences is flexible enough to include additional design information, such as the motor and propeller parameters. We then show how our novel design embedding is effective when combined with a transformer model for predicting design performance. As a result, we demonstrate that using transformers as surrogate models for design, enables faster performance evaluations and therefore more opportunity to explore a larger variation of diverse designs. 

\section{Towards Practical UAV design}

UAVs are an exciting technology that will pave the way for many unforeseen useful future use-cases. Even with today's technology, we already see a significant variation in UAV designs from food delivery/logistics \cite{gu2020vehicle}, to the detection of sharks along highly populated beaches \cite{kiszka2016using}, to air taxis. As a result, one single optimized design will not be enough for all applications. Furthermore, for each application a designer will be required to go through the same design process, but with a different set of objectives. Unfortunately, the design process of UAVs is complex and many of the design choices are highly coupled resulting in non-linear relationships. One example of such a design choice is in the selection of the motor, propeller, battery combination. All three components are intrinsically linked and therefore it is not necessarily clear in which order to select these components (see our previous work \cite{cobb2021physical}). Such coupled relationships make UAV design especially challenging, even for domain experts.

CAD plays an important role in the design of cyberphysical systems, and for UAVs this is no different. A large part of the design process is not only making important design (topological/parametric) decisions, but is also simulating designs. Once an agent, generally a human, uses their experience to propose a design, the next stage is to simulate the performance using flight dynamics models \cite{Bapty2022design, walker2022flight}. Depending on the fidelity of such models and the accuracy desired from the designer, such models can take anything from a few minutes to a few days to evaluate. Finally, once a designer is happy with their design, they must then go through the manufacturing process, prototype, do real world safety checks etc. The focus of our work is on the CAD component of the design process. Automating the evaluation of each proposed design by avoiding the use of high-fidelity models is of high value. In fact machine learning models even allow for parallelization where one can pass a batch of designs into a machine learning model in order to evaluate their performance in a matter of seconds. Replacing such simulators with machine learning models is often referred to as surrogate modeling. 

\begin{figure}[h]
\centering
\includegraphics[width=\columnwidth]{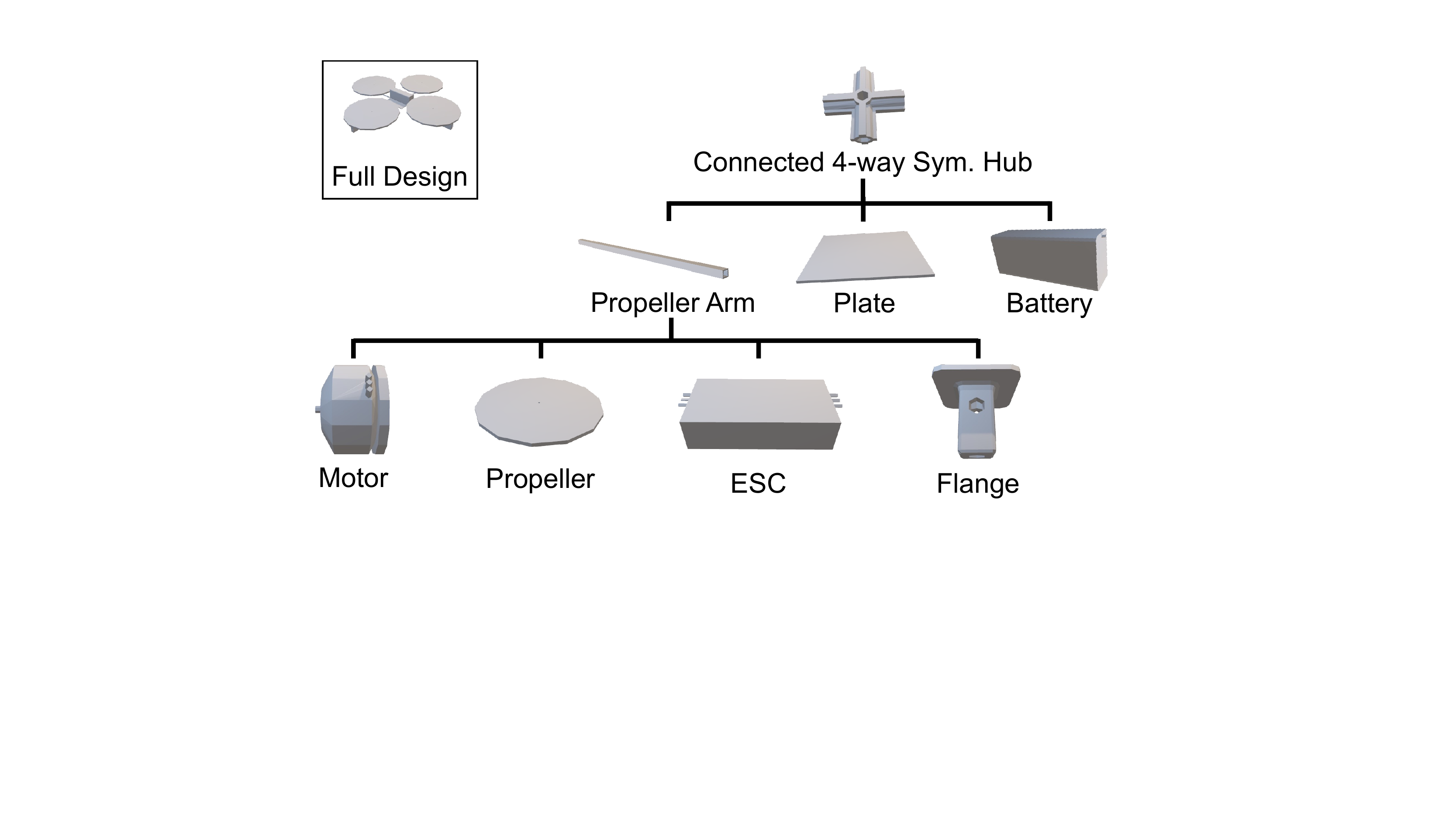}
\caption{A pictorial example of our UAV tree representation of a symmetric quadcopter. Since the design is symmetric only one propeller arm needs to be defined, which is then repeated four times when compiled to the full model as shown in the top left.}\label{fig:tree}
\end{figure}

\begin{figure}[h]
\centering
\includegraphics[width=\columnwidth]{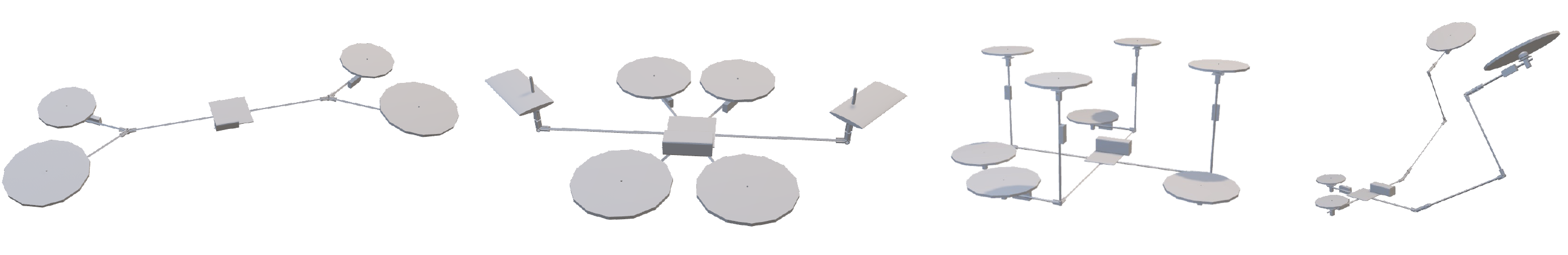}
\caption{A sample of UAV designs that we are able to build with our grammar to demonstrate the representational power of our design language.}\label{fig:uavs}
\end{figure}

One reason why the path forward to automating the design of UAVs is now achievable is due to the recent success of  deep learning models for structured data. Unlike for fully-connected neural networks and convolutional neural networks, models such as transformers \cite{vaswani2017attention} and graph neural networks \cite{wu2020comprehensive} can ingest highly structured data. This data format is more suitable to designing cyberphysical systems such as UAVs. In our work, we use a design language where UAV designs naturally follow a tree structure. This is a structure that is also prevalent for other cyberphysical systems such as in robotics \cite{zhao2020robogrammar}. As an example, Figure \ref{fig:tree} shows how a quadcopter (4-propeller UAV) can be represented as a tree, with the full design being shown in the top left corner. While this is a simple example, we are able to represent a huge diversity of designs with this tree representation as demonstrated in Figure \ref{fig:uavs}. 

\section{Representation of UAV Designs for AI Consumption}

While the tree representation outlined in the previous section provides an initial level of abstraction, a further level is required before being able to pass a full UAV design through a machine learning model. In designing an appropriate embedding we have two key objectives:
\begin{itemize}
    \item We want to work with sequences to leverage the performance of sequence models.
    \item The sequences must encode metadata of design components such as motor, propeller, and battery properties.
\end{itemize}
Following these two objectives, the first step in achieving our desired embedding is to convert the tree representation into a sequence. The tree to sequence conversion is a preorder traversal that corresponds to a flattening of the tree. In the process of flattening the tree, the parameters of each component are also included. Their order is predetermined via the design language such that we know exactly which arguments are to be expected for each type of component. For example in the conversion of the tree in Figure \ref{fig:tree}, we get the sequence as shown in Figure \ref{fig:seq}. Within this sequence, we see an example of how the arguments of each component can be flattened. First, we use key-value pairs to provide context for each value. For example, the float `\texttt{0.0}' has the key `\texttt{angle}' for context. Second, given the design language we know the order in which to expect certain key-value pairs. For example, the key-value pair `\texttt{\{`node\_type': `PropArm'\}}' means we expect to see the arm length parameter, followed by the motor, propeller, and ESC (electronic speed control) respectively. This predetermined ordering of the hierarchy means that parsing between trees and sequences is simple. Finally, our symbolic representation of design also enables exploitation of symmetry to compress design description. 
For example, if we want all four propeller arms to contain the same subsystem, then we only need to specify it once (as a single child in the design tree) and use a symmetry tag over the hub, for example, `ConnectedHub4\_Sym' denotes a hub with four connections - each having the same subsystem. 
The design assumes that the same subsystem will connect to all the arms.
\begin{figure}[h!]
\centering
\framebox[0.97\columnwidth]{%
\begin{minipage}[t]{0.8\columnwidth}
  \scriptsize
  \texttt{
  \shortstack[l]{[\\ \{`node\_type': `ConnectedHub4\_Sym'\},\\
    \{`node\_type': `PropArm'\},\\
    \{`armLength': 210.88760375976562\},\\
     \{`motorType': `t\_motor\_MN2212KV780'\},\\
     \{`propType': `apc\_propellers\_12x5'\},\\
     \{`escType': `t\_motor\_T\_80A'\},\\
     \{`offset': -3.2862548828125\},\\
     \{`offset': 4.2498626708984375\},\\
     \{`angle': 0.0\},\\
     \{`x1\_offset': 4.219192504882812\},\\
     \{`z1\_offset': 3.637290954589844\},\\
     \{`batteryType': `TurnigyGraphene1400mAh3S75C'\}\\ ]}}
\end{minipage}
}
\caption{The sequence representation of a symmetric quadcopter, such as the one in Figure \ref{fig:tree}. We build this sequence by following a preorder traversal of the design tree and ensure that the parameters of each component are included in accordance to the grammar.}\label{fig:seq}
\end{figure}

\begin{figure}[h]
\centering
\includegraphics[width=\columnwidth]{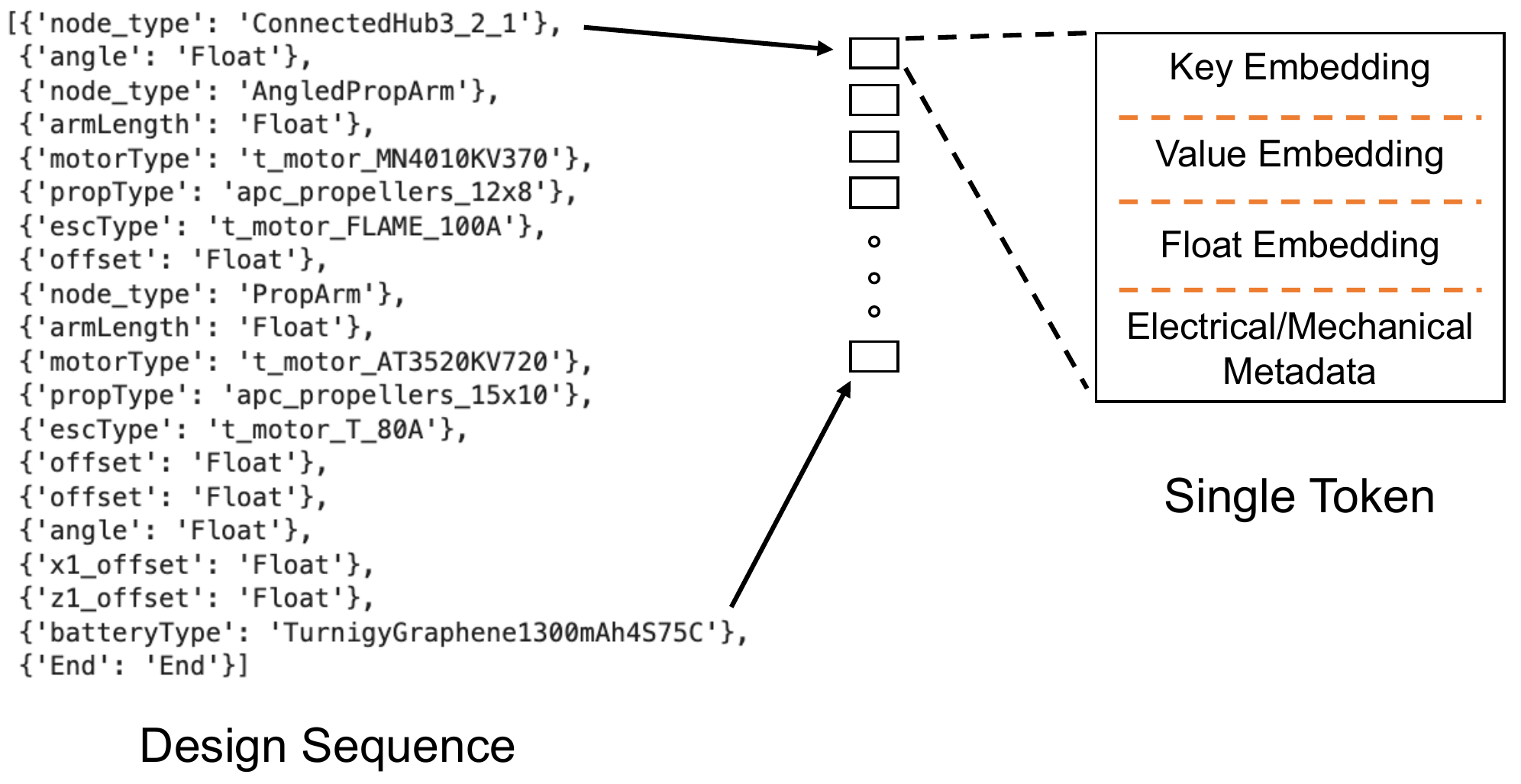}
\caption{Shows how each token in the design sequence is embedded into a vector for consumption via the machine learning model. The key and value embeddings follow a one-hot encoding, and the electrical/mechanical metadata is included according to the component being tokenized.}\label{fig:emb}
\end{figure}

Our second objective is to encode metadata such as motor, propeller, and battery properties. This means that that a machine learning model is expected to have meaning associated with an input such as \texttt{\{`motorType': `t\_motor\_MN2212KV780'\}}. As a result, when we convert each key-value pair into a token embedding, we encode any associated metadata into the token as well. Overall each token of the sequence contains a one-hot encoding for the 18 classes of keys (e.g. \texttt{motorType}, \texttt{propType}, etc.) and a one-hot encoding for the 671 classes of values (e.g. \texttt{0.0}, \texttt{t\_motor\_T\_80A}, etc.), as well as 51 possible attributes of electrical/mechanical components, and a final dimension for float values. Figure \ref{fig:emb} provides an overview of this conversion.

\section{UAV Procedural Generator}

In the previous two sections we defined a tree representation and a corresponding procedure to convert from the tree to a sequence. An important and necessary advantage of defining a logical UAV grammar, is that we can build a procedural generator that generates UAV design trees with minimal compute. In fact, we can sample from our generator, by sampling its stochastic parameters to draw hundreds of thousands of designs in the order of minutes. Therefore, design generation becomes trivial, but design evaluation remains a challenge due to the computational cost of running the UAV simulation software.

We note that defining a language for each cyberphysical system requires significant domain expertise. A procedural generator requires a suitable design representation, e.g. a tree, and a broad enough vocabulary to capture a large diversity of designs. In our development of our procedural generator (or declarative probabilistic program), we started by initially generating variants of the quadcopter (see Figure \ref{fig:tree}) and iterated towards more complex designs that included wings and a multitude of different component combinations. In summary, our procedural generator can generate a variety of designs that are topologically valid. By \textit{topologically valid}, we mean that they can be evaluated by the available scientific models. However, they may not be structurally sound or able to fly. In the next section we demonstrate how machine learning approaches can aid in producing working UAV designs.

\section{Transformers as Surrogate Models for UAV Design}

As alluded to earlier, we are focused on speeding up the computational stage of the UAV design process. Therefore, rather than relying on expensive flight simulation models, our aim is to reduce the cost of computation time to the order of seconds per design. Reducing this computational cost will facilitate faster design exploration and will therefore open up the UAV design domain to further machine learning exploratory approaches.  

\subsection{Flight Dynamics Model}

The performance of each UAV is assessed using a pipeline comprising CAD tools such as Creo~\cite{creo} and a custom flight dynamics model (FDM)~\cite{walker2022flight,Bapty2022design}. Each UAV is assessed on controllability (existence of trim states) at different speeds. In particular, the FDM evaluates whether the UAV can fly at a specific velocity by adjusting the controls and orientation of a vehicle until the state of the vehicle reaches the desired value. These adjustments are achieved numerically, using the MINPACK package and the nonlinear simplex algorithm (see \citet{walker2022flight} for more details). Before evaluating the design in the FDM, each design must also be compiled and evaluated in a solid modeling computer-aided design tool, such as creo. This tool provides information to the FDM such as the overall mass, the moments of inertia, and potential interferences between parts. The output of the FDM provides a range of UAV statistics, such as trim states and electrical performance (power, current, voltage etc.).

\subsection{Objective}

As design is an iterative process, and we have the capability of producing hundreds of thousands of topologically valid designs, one might want to initially filter through these designs (quickly) to find a subset that meet designer specifications. A first step in this process is to find UAV designs that are able to fly or hover. As a result, we define the first objective as one where we aim to predict whether a design can hover, as this is often a vital characteristic of UAVs.\footnote{We note that some UAVs, such as many fixed wing UAVs, are launched (or catapulted) and therefore are not required to hover. For this class of UAVs, we would choose a different objective for filtering.} We therefore use a binary cross-entropy loss as the objective, whereby we label a design as $y = 1$ for hover times greater than $0$ and $y = 0$ for hover times of $0$.

\subsection{Transformer Model}

We will now introduce our transformer model that will operate over the sequence of token embeddings as highlighted in Figure \ref{fig:emb}. One key innovation of modern deep learning approaches is in the ability of some architectures to ingest structured data. In this paper so far, we have shown how we were able to define a flexible UAV grammar that allows us to represent a large diversity of designs as sequences. Transformers \cite{vaswani2017attention} are well-known to be extremely effective at operating on sequence data. They have seen huge success as part of large language models \cite{devlin2018bert, brown2020language} and we will now show how our UAV design embedding can bring success to the cyberphysical domain. 

In order to classify whether a design can hover, we build a transformer encoder that consists of a linear encoding layer followed by a positional encoding layer. The output of the positional encoding is then passed through a transformer encoding layer (\texttt{nn.TransformerEncoder} in PyTorch) with 8 layers and 2 heads. The final token from the transformer encoder is then passed through a linear layer that goes from the embedding dimension of 200 to a single output dimension. During training we use the PyTorch in-built SGD optimizer with a learning rate of 0.01 and set the number of epochs to 2500. 

\subsection{Initial Data and Results}

\begin{figure}[h!]
\centering
    \includegraphics[width=\columnwidth]{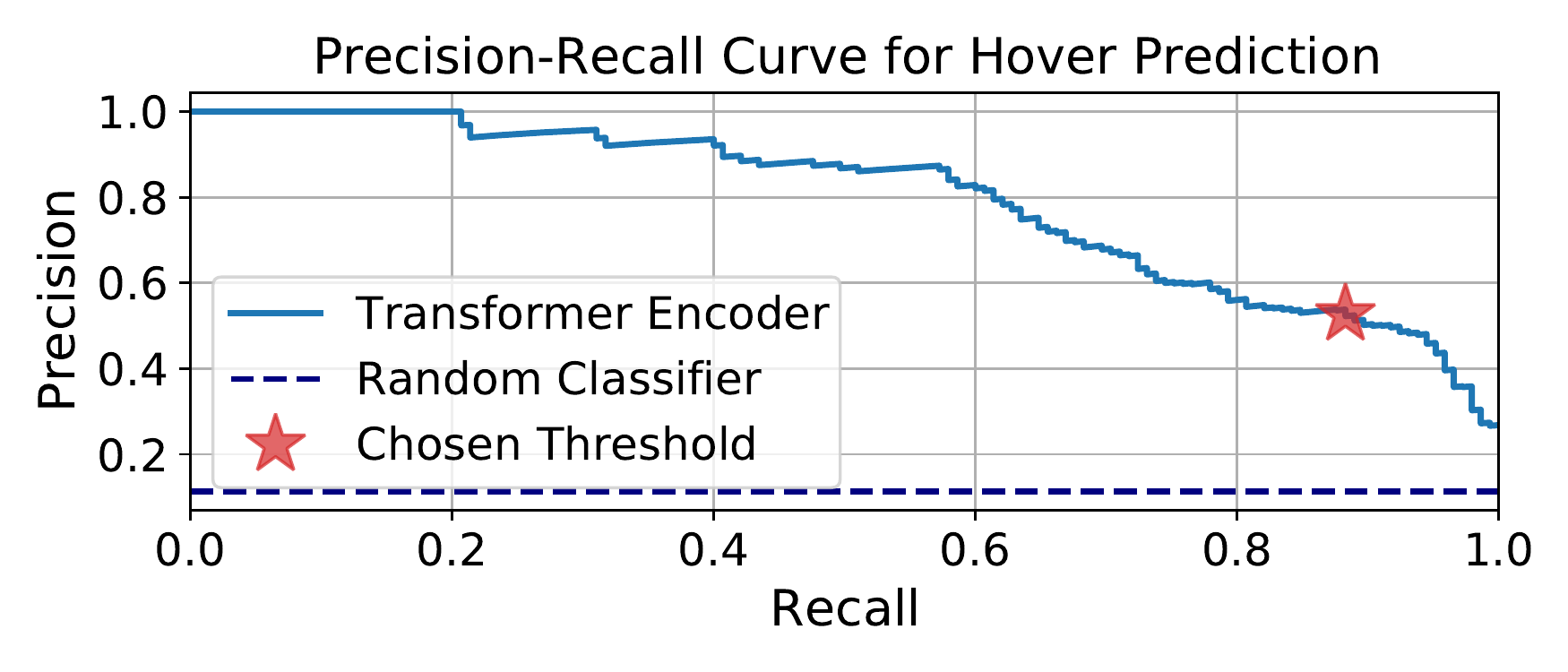}
    \caption{Precision-recall curve for the transformer encoder. The graph highlights the relative performance compared to a random classifier and indicates our chosen precision-recall threshold with the red star. } 
    \label{fig:PRCurve}
\end{figure}

For the initial training of our model, we require a labeled data set. We therefore run the full CAD pipeline for $6{,}352$ UAV designs, where each design is sampled randomly from the procedural generator. Within this data set, only $794$ UAVs have a hover time greater than $0$~s. While computationally expensive, this initial data set is large enough to train our transformer model. We set aside $80~\%$ of the data for training and $20~\%$ for test. The resulting performance of our transformer model is an accuracy of $93.6~\%$. However, at this accuracy the recall for hovering UAVs is $0.63$, which means that we would miss around $40 \%$ of designs with this desired characteristic in the downstream tasks. Therefore, with careful calibration by changing the classification threshold from $0.5$ to $0.15$, we can achieve a recall of $0.88$ at the expense of a precision of $0.53$. When we describe the pipeline in the next section, a threshold of $0.15$ will result in the collection of a balanced data set of diverse hovering UAVs. Figure \ref{fig:PRCurve} provides further context for our threshold choice. The plot shows the precision-recall curve for the transformer encoder, where we have highlighted our chosen precesion-recall threshold with a red star. Depending on the recall, precision, and available compute, one could select a threshold accordingly by analyzing this graph.

\subsection{Design Pipeline and Experimental Results}

\begin{figure}[h]
\centering
\includegraphics[width=\columnwidth]{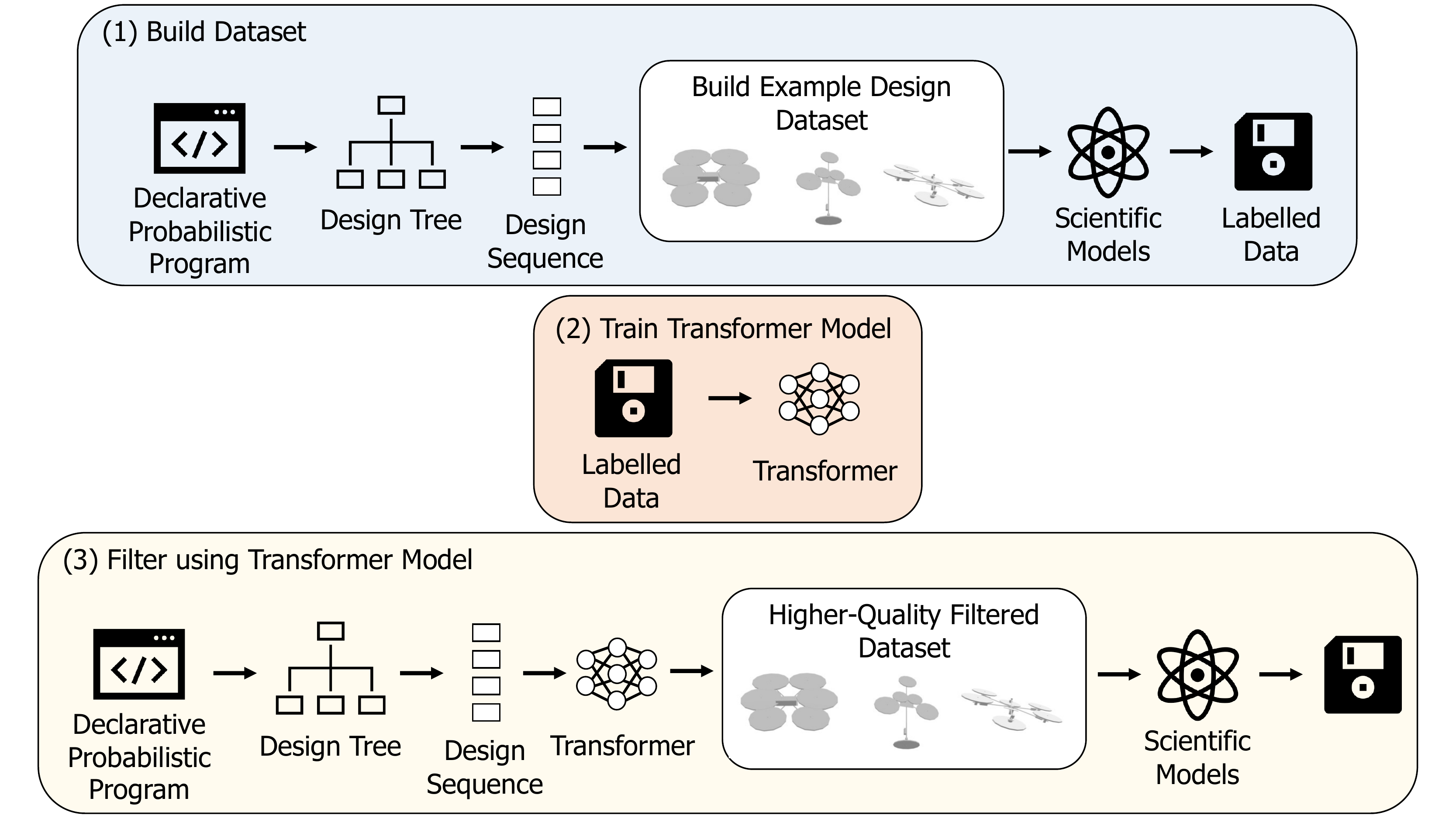}
\caption{Graphic to provide an overview of the UAV design pipeline. The initial stage is to build a data set by sampling from the stochastic procedural generator (the declarative probabilistic program) in order to build a data set via running these designs through the scientific flight dynamics models. The next stage is to train the transformer on this data. In the last stage, we use the trained transformer to perform rejection sampling over the procedural generator to help find higher-quality designs.}\label{fig:pipeline}
\end{figure}

We can now describe the entire design pipeline and our bootstrapping method for jointly improving overall UAV design and our surrogate models. Figure \ref{fig:pipeline} outlines the three staged process. In stage 1, highlighted in blue, we collect the initial labeled data. This requires sampling trees from our procedural generator and then flattening these trees into the appropriate format for the scientific models. We then run the scientific models (creo and the FDM) to evaluate the performance to build the initial data set. In stage 2, highlighted in orange, we train our transformer model as described in the previous section. Finally in stage 3, highlighted in yellow, we use our transformer model to filter through designs as sampled from the procedural generator in order to build a higher quality data set.

To demonstrate our pipeline, we sample $100{,}000$ new designs from the procedural generator and then use the transformer encoder with a threshold of $0.15$ to filter out designs that are predicted to be unable to hover. The result is $21{,}800$ UAV designs. This result is consistent with the performance over the validation data which reported an expected precision of around $50~\%$. Given that randomly sampling from the generator provides roughly $10~\%$ of designs that can hover and our model has a precision of $0.53$ and a recall of 0.88, we would expect to be left with around $20{,}000$ UAV designs of which $10{,}000$ should be able to hover. This result is significant as each evaluation through the scientific models takes 4-10 minutes depending on the complexity of the design. We can estimate a lower bound on the compute time by assuming 4 minutes per FDM evaluation, then $100{,}000$ UAVs would take $277.7$ days of compute, compared to $60.6$ days. 

\begin{figure}
     \centering
     \begin{subfigure}[b]{0.9\columnwidth}
         \centering
         \includegraphics[width=\textwidth]{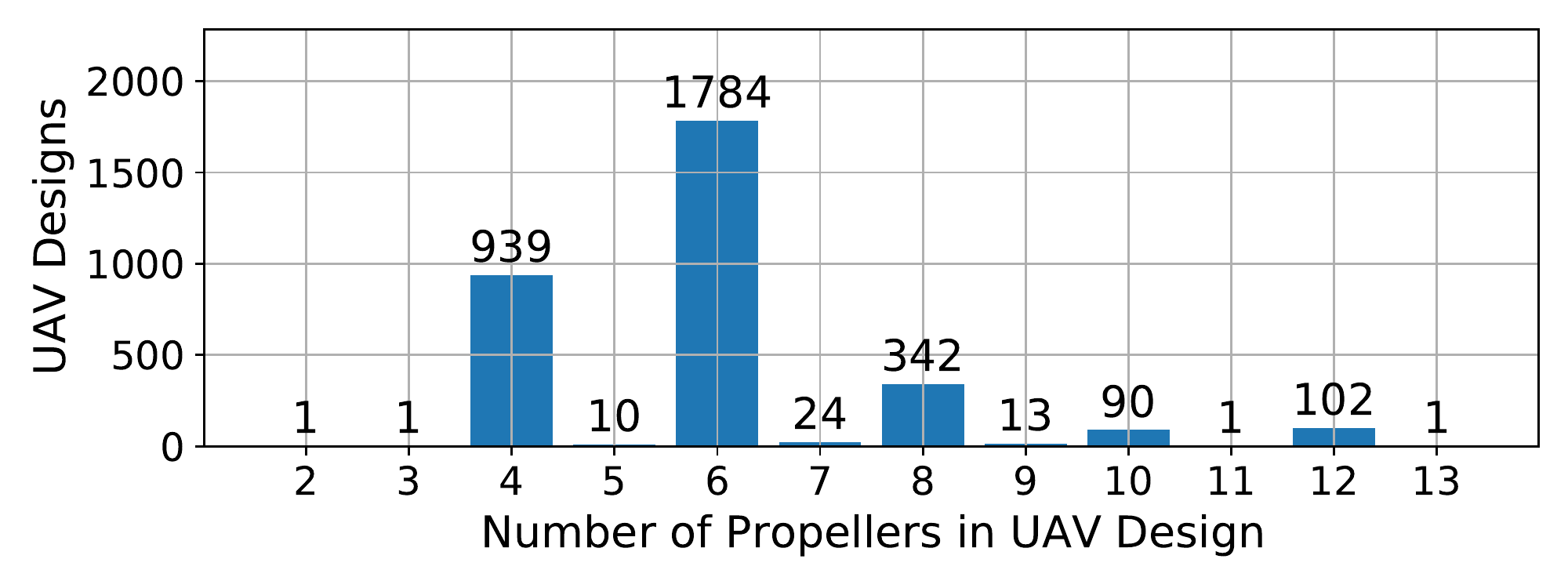}
         \caption{\# Propellers in UAV designs that can hover.}
         \label{fig:props}
     \end{subfigure}
     \hfill
     \begin{subfigure}[b]{0.9\columnwidth}
         \centering
         \includegraphics[width=\textwidth]{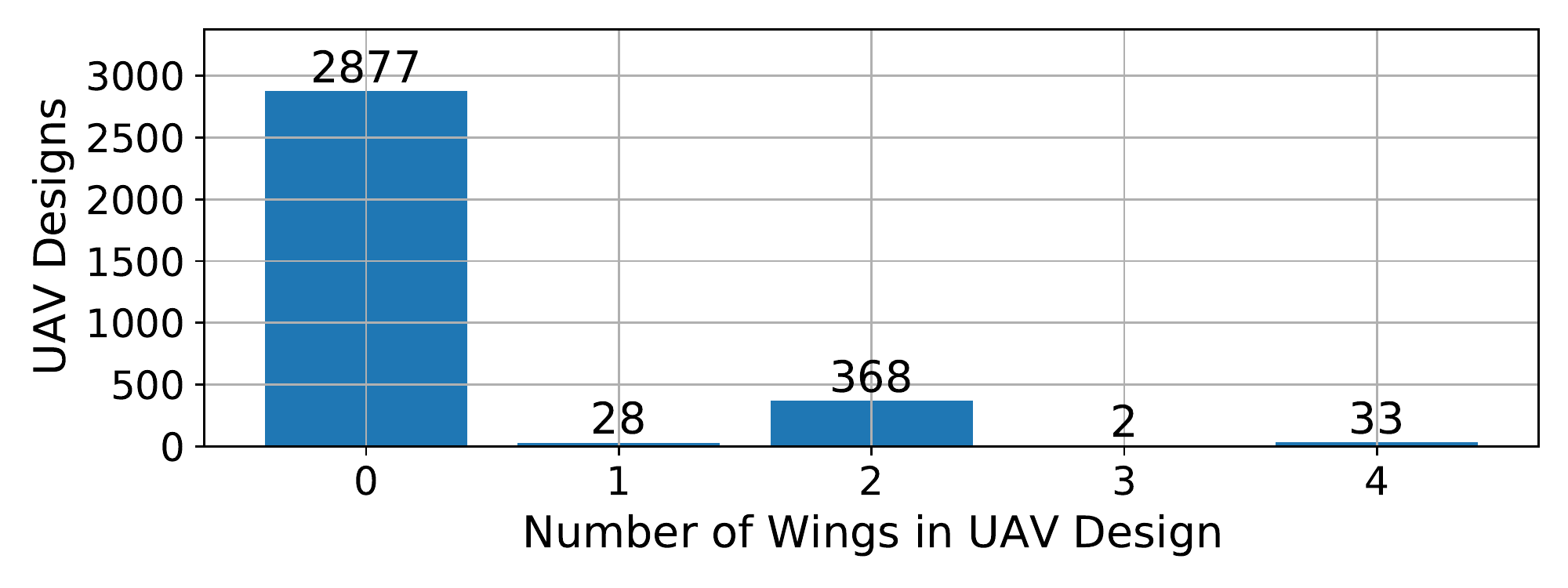}
         \caption{\# Wings in UAV designs that can hover.}
         \label{fig:wings}
     \end{subfigure}
        \caption{Quantitative results of the filtering process. After using the transformer encoder, we are left with a diverse range of UAV designs that can hover. We see designs ranging from two propellers to thirteen (a) and designs with number of wings ranging from none to 4 (b).}
        \label{fig:bars}
\end{figure}

\begin{figure*}[h!]
\centering
    \includegraphics[width=0.75\textwidth]{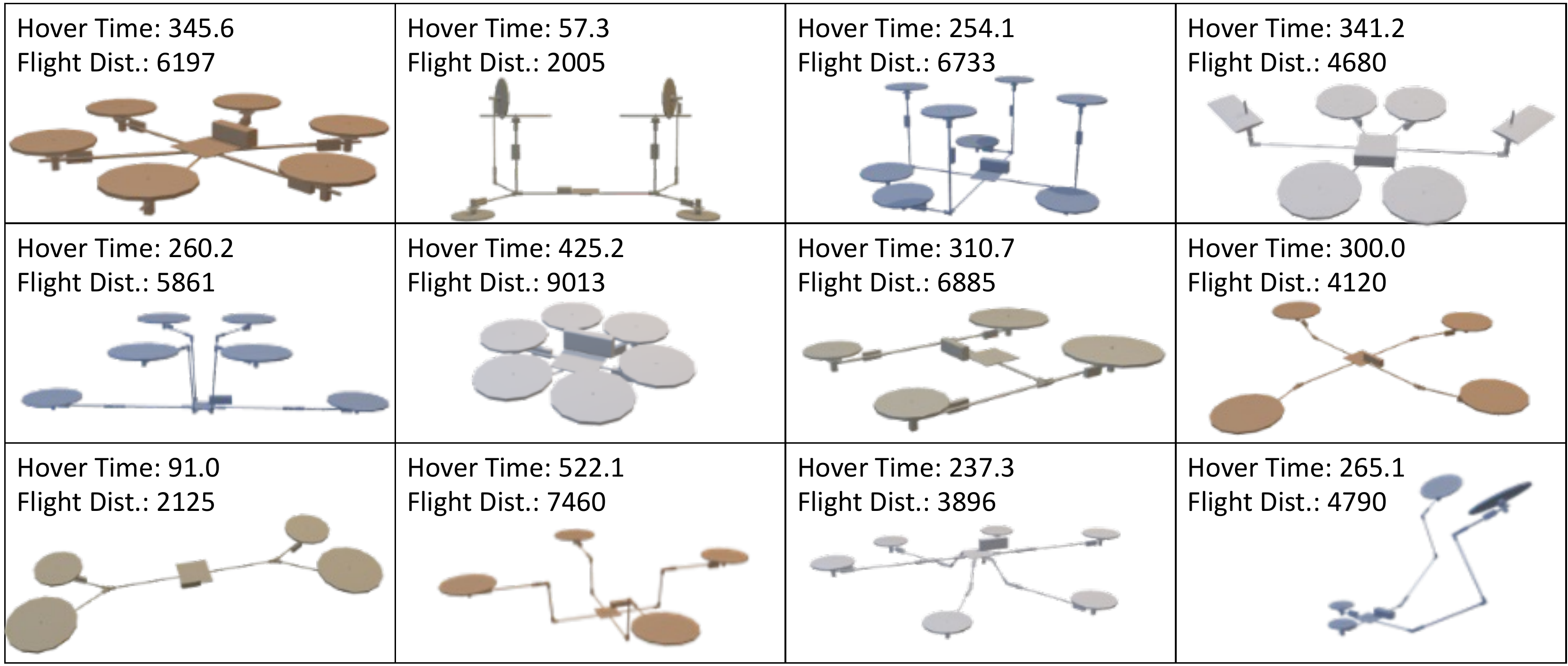}
    \caption{A subset of UAV designs that were found via the transformer-assisted design pipeline. The high proportion of UAV designs that meet the objective of being able to hover are also diverse. Many designs are novel and do not follow the standard choices that an expert UAV designer would necessarily select. } 
    \label{fig:UAVExamples}
\end{figure*}

To determine the success of our filtering approach, we evaluated a subset of $6{,}621$ UAV designs (out of the proposed $21{,}800$) using the scientific models. It is at this point where we see the generalization of our approach. Of these $6{,}621$ designs, $3{,}308$ met the desired objective of being able to hover. This new data set therefore consists of $50 \%$ of UAVs that have the ability to hover compared to the $12.5 \%$ from the previous iteration (stage 1 in Figure \ref{fig:pipeline}). This result is consistent with the performance over the validation set, where we also saw a comparable precision of $0.53$.
One potential concern is that the transformer encoder could identify UAV designs that are less diverse. However we show with both quantitative and qualitative results that a lack of diversity is not an issue. For quantitative results we refer to Figure \ref{fig:bars}. Figure \ref{fig:bars} presents the distribution of propellers (\ref{fig:props}) and wings (\ref{fig:wings}) for all hovering UAV designs that were filtered out via stage 3. We can see that the designs retrieved using our transformer model have a large range of values for both number of propellers and number of wings. Notably, we see from Figure \ref{fig:props} that a large proportion of the UAVs that hover have an even number of propellers. This observation meets what we see in practice with many UAVs taking on the form of quadcopters and hexcopters. However, this design process also provided some novel designs such as a 13 propeller UAV (or \textit{``tridecacopter''}) with a hover time of $204$~s. We see a similar pattern in Figure \ref{fig:wings} where there are more hovering UAVs with an even number of wings than odd. We further note that the vast majority of hovering UAV designs do not have wings. We postulate that the objective requirement of just being able to hover does not favor the inclusion of wings compared to other objectives such as maximum distance. For qualitative results, we select a range of filtered designs and present them in Figure \ref{fig:UAVExamples}. Importantly the UAVs in this figure highlight the diversity in designs that were retrieved via our AI-assisted UAV design process. These designs all meet the criteria of being able to hover, as well as having some other favorable performance metrics that were not previously specified such as reasonable flight distances. While some of these designs conceivably could have been designed by domain experts, such as the hexcopters, many designs would not have been likely expert choices. Therefore, our AI-assisted design process could lead to many more interesting new designs that may not have been considered before.

We now ask how does this new data set help with the design process and what does it mean for the future of UAV design and the design of cyberphysical systems in general. The designs in Figure \ref{fig:UAVExamples} would not have been realistically possible to compute without the use of AI, namely the transformer encoder. The ability of the transformer to accurately predict whether a design would be able to hover without the need of running the scientific models is a big step in the process of automating the UAV design process. The transformer reduces the $1$ in $10$ chance of evaluating a hovering design to $1$ in $2$, and saves significant compute time. As a result of this saving, we enable the possibility of evaluating many more interesting novel designs, such as those shown in the Figure. For future UAV design our process as described in this paper has the potential to facilitate even more sophisticated machine learning approaches that are able to leverage the higher quality data sets that our approach produces. More generally, our case study demonstrates how a cyberphysical system such as a UAV can be represented as a tree structure using a domain-specific design language. In highlighting this demonstration, we believe other cyberphysical systems could be represented in flexible grammars that could then utilize sequence-based transformer approaches as shown here.

\section{Related Work}

The use of machine learning in computer aided design (CAD) has gained significant attention, and only a few works have been proposed in recent literature that develop machine learning approaches. SketchGraphs dataset~\cite{seff2020sketchgraphs} is a collection of sketches extracted from parametric CAD models which begin as two-dimensional (2D) sketches consisting of geometric primitives (e.g., line segments, arcs) and explicit constraints between them (e.g., coincidence, perpendicularity) that form the basis for three-dimensional (3D) construction operations. This dataset has been used for generative model of CAD sketches~\cite{willis2021engineering}, and other applications of learning in physical design~\cite{seff2021vitruvion, para2021sketchgen}.
Building a design grammar has been the subject of other works such as in \citet{zhao2020robogrammar}. There are also many other works that have focused on specific parts such as robot arms \cite{xu2021end} and airfoil design \cite{chen2022learning}. Two key differences between our approach to design and these previous works is: (1) Our focus on a pathway towards deployment by using physically realizable components; (2) The use of transformer models on our design embeddings to directly evaluate performance.

\section{Future Deployment}

All the electrical and mechanical components used in our UAV design pipeline are readily available UAV parts that exist on the market. Therefore, while we are yet to manufacture and deploy physical models of these designs in the real world, we believe that this is the obvious next step in the path towards full deployment. We see two challenges associated with this next step of achieving full deployment. The first challenge is taking a design proposed by our AI-assisted pipeline and ensuring that it is physically possible to build. It is likely that some designs deemed valid according to the high-fidelity simulator may in fact not be feasible designs. This potential inconsistency between the flight simulator and the real-world behavior is often an inevitable part of the prototyping process. 
Overcoming this challenge will require human input on our most promising designs to ensure that the location of components, such as electrical, are all able to be connected in a way that is physically realizable. We believe that the fact that we are using real parts and account for interferences between components means that we have mitigated against this challenge to a large degree, however we expect human input to be important at this stage. Finally, the second challenge is the need for a formal way of incorporating more human input. This challenge, which is also linked to the first challenge, will be key going forward for almost all AI-assisted CAD tools across multiple domains. In its current form, our pipeline represents an important and novel first step towards using AI to reduce the computational cost of UAV design. As a result, our AI-assisted pipeline can guide the designer into regions of the design space that they may never have thought of before. We are actively looking into incorporating new objective functions for our transformer model that better suit designer preferences, as well as using new models to better incorporate feedback. 

\section{Conclusion}

The application of AI for the computational design of cyberphysical systems is an emerging domain where the gap between basic AI research and real deployment is becoming narrower. In this paper, we have demonstrated that the use of machine learning approaches can significantly speed up the design process for UAVs. Specifically, in using transformer encoder models trained on UAV design sequences, we have shown the utility of sequence models for design. The novelty of this work is both in our definition of the UAV design language and our deployment of a transformer model. Our UAV design language can represent a huge diversity of designs in a way that can then be tokenized for use in sequence models. Our approach reduced the computational cost of evaluating $100{,}000$ UAV designs from $277.7$ days of compute to $60.6$ days by early identification and removal of poorly performing UAV designs. Finally, our AI-assisted design pipeline led to a large diversity of high-performing UAV designs, of which many do not follow the standard choices that an expert would select (e.g. see Figure \ref{fig:UAVExamples}). As a result, our approach to UAV design opens up the possibility of using AI-tools for developing new designs that could be more efficient and better suited to future task-specific applications.


\section*{Acknowledgments}
This project was supported by DARPA under the Symbiotic
Design for Cyber-Physical Systems (SDCPS) with contract 
FA8750-20-C-0002. 
The views, opinions and/or findings expressed
are those of the author and should not be interpreted as
representing the official views or policies of the Department
of Defense or the U.S. Government.

\bibliography{aaai23}

\end{document}